\documentclass[submission,copyright,creativecommons]{eptcs}
\providecommand{\event}{E2ECMVCS 2022} 

\title{Assessing the Unitary RNN as an End-to-End Compositional Model of Syntax}
\author{
Jean-Philippe Bernardy \qquad\qquad Shalom Lappin
\institute{  Centre for Linguistic Theory and Studies in Probability  \\
  Department of Philosophy, Linguistics and Theory of Science\\
  University of Gothenburg}
\email{  jean-philippe.bernardy@gu.se \qquad\qquad  shalom.lappin@gu.se  }
}

\def\titlerunning{The Unitary RNN as an End-to-End Compositional Model of Syntax}
\def\authorrunning{Jean-Philippe Bernardy and Shalom Lappin}

\usepackage{pgfplots}
\pgfplotsset{cycle list name=color list}

\usepackage{tikz}
\usetikzlibrary{calc}
\usetikzlibrary{positioning, fit, arrows.meta, shapes}
\usepackage{enumitem}
\newlist{lingex}{enumerate}{3} 
\setlist[lingex,1]{parsep=0pt,itemsep=1pt,label=(\Alph*),resume=lingexcount}

\usepackage{newunicodechar}
\newunicodechar{ }{\,}
\newunicodechar{¬}{\ensuremath{\neg}} %
\newunicodechar{²}{^2}
\newunicodechar{³}{^3}
\newunicodechar{·}{\ensuremath{\cdot}}
\newunicodechar{¹}{^1}
\newunicodechar{×}{\ensuremath{\times}} %
\newunicodechar{÷}{\ensuremath{\div}} %
\newunicodechar{ʲ}{^j}
\newunicodechar{ʳ}{^r}
\newunicodechar{ˡ}{^l}
\newunicodechar{̷}{\not} %
\newunicodechar{Γ}{\ensuremath{\Gamma}}   %
\newunicodechar{Δ}{\ensuremath{\Delta}} %
\newunicodechar{Η}{\ensuremath{\textrm{H}}} %
\newunicodechar{Θ}{\ensuremath{\Theta}} %
\newunicodechar{Λ}{\ensuremath{\Lambda}} %
\newunicodechar{Ξ}{\ensuremath{\Xi}} %
\newunicodechar{Π}{\ensuremath{\Pi}}   %
\newunicodechar{Σ}{\ensuremath{\Sigma}} %
\newunicodechar{Φ}{\ensuremath{\Phi}} %
\newunicodechar{Ψ}{\ensuremath{\Psi}} %
\newunicodechar{Ω}{\ensuremath{\Omega}} %
\newunicodechar{α}{\ensuremath{\mathnormal{\alpha}}}
\newunicodechar{β}{\ensuremath{\beta}} %
\newunicodechar{γ}{\ensuremath{\mathnormal{\gamma}}} %
\newunicodechar{δ}{\ensuremath{\mathnormal{\delta}}} %
\newunicodechar{ε}{\ensuremath{\mathnormal{\varepsilon}}} %
\newunicodechar{ζ}{\ensuremath{\mathnormal{\zeta}}} %
\newunicodechar{η}{\ensuremath{\mathnormal{\eta}}} %
\newunicodechar{θ}{\ensuremath{\mathnormal{\theta}}} %
\newunicodechar{ι}{\ensuremath{\mathnormal{\iota}}} %
\newunicodechar{κ}{\ensuremath{\mathnormal{\kappa}}} %
\newunicodechar{λ}{\ensuremath{\mathnormal{\lambda}}} %
\newunicodechar{μ}{\ensuremath{\mathnormal{\mu}}} %
\newunicodechar{ν}{\ensuremath{\mathnormal{\mu}}} %
\newunicodechar{ξ}{\ensuremath{\mathnormal{\xi}}} %
\newunicodechar{π}{\ensuremath{\mathnormal{\pi}}}
\newunicodechar{π}{\ensuremath{\mathnormal{\pi}}} %
\newunicodechar{ρ}{\ensuremath{\mathnormal{\rho}}} %
\newunicodechar{σ}{\ensuremath{\mathnormal{\sigma}}} %
\newunicodechar{τ}{\ensuremath{\mathnormal{\tau}}} %
\newunicodechar{φ}{\ensuremath{\mathnormal{\phi}}} %
\newunicodechar{χ}{\ensuremath{\mathnormal{\chi}}} %
\newunicodechar{ψ}{\ensuremath{\mathnormal{\psi}}} %
\newunicodechar{ω}{\ensuremath{\mathnormal{\omega}}} %
\newunicodechar{ϕ}{\ensuremath{\mathnormal{\phi}}} %
\newunicodechar{ϵ}{\ensuremath{\mathnormal{\epsilon}}} %
\newunicodechar{ᵀ}{\ensuremath{^T}}
\newunicodechar{ᵏ}{^k}
\newunicodechar{ᵐ}{^m}
\newunicodechar{ᵢ}{\ensuremath{_i}} %
\newunicodechar{ᶿ}{^\theta}
\newunicodechar{ }{\quad}
\newunicodechar{†}{\dagger}
\newunicodechar{ }{\,}
\newunicodechar{′}{\ensuremath{^\prime}}  %
\newunicodechar{″}{\ensuremath{^\second}} %
\newunicodechar{‴}{\ensuremath{^\third}}  %
\newunicodechar{ⁱ}{^i}
\newunicodechar{⁵}{\ensuremath{^5}}
\newunicodechar{⁺}{\ensuremath{^+}} 
\newunicodechar{⁻}{\ensuremath{^-}} 
\newunicodechar{ⁿ}{^n}
\newunicodechar{₀}{\ensuremath{_0}} %
\newunicodechar{₁}{\ensuremath{_1}} 
\newunicodechar{₂}{\ensuremath{_2}} 
\newunicodechar{₃}{\ensuremath{_3}}
\newunicodechar{₊}{\ensuremath{_+}} 
\newunicodechar{₋}{\ensuremath{_-}} 
\newunicodechar{ₘ}{_m}
\newunicodechar{ₙ}{_n} %
\newunicodechar{⃰}{\ensuremath{^{\ast}}}
\newunicodechar{ℂ}{\ensuremath{\mathbb{C}}} %
\newunicodechar{ℒ}{\ensuremath{\mathscr{L}}}
\newunicodechar{ℕ}{\mathbb{N}} %
\newunicodechar{ℚ}{\ensuremath{\mathbb{Q}}}
\newunicodechar{ℝ}{\ensuremath{\mathbb{R}}} %
\newunicodechar{ℤ}{\ensuremath{\mathbb{Z}}} %
\newunicodechar{ℳ}{\mathscr{M}}
\newunicodechar{⅋}{\ensuremath{\parr}} %
\newunicodechar{←}{\ensuremath{\leftarrow}} %
\newunicodechar{↑}{\ensuremath{\uparrow}} %
\newunicodechar{→}{\ensuremath{\rightarrow}} %
\newunicodechar{↔}{\ensuremath{\leftrightarrow}} %
\newunicodechar{↖}{\nwarrow} %
\newunicodechar{↗}{\nearrow} %
\newunicodechar{↝}{\ensuremath{\leadsto}}
\newunicodechar{↦}{\ensuremath{\mapsto}}
\newunicodechar{⇆}{\ensuremath{\leftrightarrows}} %
\newunicodechar{⇐}{\ensuremath{\Leftarrow}} %
\newunicodechar{⇒}{\ensuremath{\Rightarrow}} %
\newunicodechar{⇔}{\ensuremath{\Leftrightarrow}} %
\newunicodechar{∀}{\ensuremath{\forall}}   %
\newunicodechar{∂}{\ensuremath{\partial}}
\newunicodechar{∃}{\ensuremath{\exists}} %
\newunicodechar{∅}{\ensuremath{\varnothing}} %
\newunicodechar{∈}{\ensuremath{\in}}
\newunicodechar{∉}{\ensuremath{\not\in}} %
\newunicodechar{∋}{\ensuremath{\ni}}  %
\newunicodechar{∎}{\ensuremath{\qed}}
\newunicodechar{∏}{\prod}
\newunicodechar{∑}{\sum}
\newunicodechar{∗}{\ensuremath{\ast}} %
\newunicodechar{∘}{\ensuremath{\circ}} %
\newunicodechar{∙}{\ensuremath{\bullet}} 
\newunicodechar{∝}{\propto}
\newunicodechar{∞}{\ensuremath{\infty}} %
\newunicodechar{∣}{\ensuremath{\mid}} %
\newunicodechar{∧}{\wedge}%
\newunicodechar{∨}{\vee}%
\newunicodechar{∩}{\ensuremath{\cap}} %
\newunicodechar{∪}{\ensuremath{\cup}} %
\newunicodechar{∫}{\int}
\newunicodechar{∷}{::} %
\newunicodechar{∼}{\ensuremath{\sim}} %
\newunicodechar{≃}{\ensuremath{\simeq}} %
\newunicodechar{≅}{\ensuremath{\cong}} %
\newunicodechar{≈}{\ensuremath{\approx}} %
\newunicodechar{≜}{\ensuremath{\stackrel{\scriptscriptstyle {\triangle}}{=}}} 
\newunicodechar{≝}{\ensuremath{\stackrel{\scriptscriptstyle {\text{def}}}{=}}}
\newunicodechar{≟}{\ensuremath{\stackrel {_\text{\textbf{?}}}{\text{\textbf{=}}\negthickspace\negthickspace\text{\textbf{=}}}}} 
\newunicodechar{≠}{\ensuremath{\neq}}%
\newunicodechar{≡}{\ensuremath{\equiv}}%
\newunicodechar{≤}{\ensuremath{\le}} %
\newunicodechar{≥}{\ensuremath{\ge}} %
\newunicodechar{⊂}{\ensuremath{\subset}} %
\newunicodechar{⊃}{\ensuremath{\supset}} %
\newunicodechar{⊆}{\ensuremath{\subseteq}} %
\newunicodechar{⊇}{\ensuremath{\supseteq}} %
\newunicodechar{⊎}{\ensuremath{\uplus}} %
\newunicodechar{⊑}{\ensuremath{\sqsubseteq}} %
\newunicodechar{⊒}{\ensuremath{\sqsupseteq}} %
\newunicodechar{⊓}{\ensuremath{\sqcap}} %
\newunicodechar{⊔}{\ensuremath{\sqcup}} %
\newunicodechar{⊕}{\ensuremath{\oplus}} %
\newunicodechar{⊗}{\ensuremath{\otimes}} %
\newunicodechar{⊙}{\ensuremath{\odot}} %
\newunicodechar{⊛}{\ensuremath{\circledast}}
\newunicodechar{⊢}{\ensuremath{\vdash}} %
\newunicodechar{⊤}{\ensuremath{\top}}
\newunicodechar{⊥}{\ensuremath{\bot}} 
\newunicodechar{⊧}{\models} %
\newunicodechar{⊨}{\models} %
\newunicodechar{⊩}{\Vdash}
\newunicodechar{⊬}{\not \vdash}
\newunicodechar{⊸}{\ensuremath{\multimap}} %
\newunicodechar{⋁}{\ensuremath{\bigvee}}
\newunicodechar{⋃}{\ensuremath{\bigcup}} %
\newunicodechar{⋄}{\ensuremath{\diamond}} %
\newunicodechar{⋅}{\ensuremath{\cdot}}
\newunicodechar{⋆}{\ensuremath{\star}} %
\newunicodechar{⋮}{\ensuremath{\vdots}} %
\newunicodechar{⋯}{\ensuremath{\cdots}} %
\newunicodechar{─}{---}
\newunicodechar{■}{\ensuremath{\blacksquare}} 
\newunicodechar{□}{\ensuremath{\square}} %
\newunicodechar{▴}{\ensuremath{\blacktriangledown}}
\newunicodechar{▵}{\ensuremath{\triangle}}
\newunicodechar{▹}{\ensuremath{\triangleright}} %
\newunicodechar{▻}{\ensuremath{\rhd}} %
\newunicodechar{▾}{\ensuremath{\blacktriangle}}
\newunicodechar{▿}{\ensuremath{\triangledown}}
\newunicodechar{◃}{\ensuremath{\triangleleft}}
\newunicodechar{◅}{\ensuremath{\lhd}}
\newunicodechar{◇}{\ensuremath{\diamond}} %
\newunicodechar{◽}{\ensuremath{\square}}
\newunicodechar{★}{\ensuremath{\star}}   %
\newunicodechar{♭}{\ensuremath{\flat}} %
\newunicodechar{♯}{\ensuremath{\sharp}} %
\newunicodechar{✓}{\ensuremath{\checkmark}} %
\newunicodechar{⟂}{\ensuremath{^\bot}} 
\newunicodechar{⟦}{\ensuremath{\llbracket}}
\newunicodechar{⟧}{\ensuremath{\rrbracket}}
\newunicodechar{⟨}{\ensuremath{\langle}} %
\newunicodechar{⟩}{\ensuremath{\rangle}} %
\newunicodechar{⟶}{{\longrightarrow}}
\newunicodechar{⟷} {\ensuremath{\leftrightarrow}}
\newunicodechar{⟹}{\ensuremath{\Longrightarrow}} %
\newunicodechar{⦇}{\rlparenthesis}
\newunicodechar{⦈}{\llparenthesis}
\newunicodechar{⪆}{\ensuremath{\gtrapprox}}
\newunicodechar{⪅}{\ensuremath{\lessapprox}}
\newunicodechar{ⱼ}{_j} %
\newunicodechar{𝒟}{\ensuremath{\mathcal{D}}} %
\newunicodechar{𝒢}{\ensuremath{\mathcal{G}}}
\newunicodechar{𝒦}{\ensuremath{\mathcal{K}}} %
\newunicodechar{𝒫}{\ensuremath{\mathcal{P}}}
\newunicodechar{𝔸}{\ensuremath{\mathds{A}}} %
\newunicodechar{𝔹}{\ensuremath{\mathds{B}}} %
\newunicodechar{𝔼}{\mathds{E}}
\newunicodechar{𝟙}{\ensuremath{\mathds{1}}} %

\newcommand\braces[1]{\{#1\}}
\newcommand\Lang{\mathcal C}
\usepackage{amsmath}
\usepackage{amssymb}
\usepackage{stmaryrd}
\usepackage{subcaption}
\usepackage[capitalize]{cleveref}
\usepackage[textsize=tiny]{todonotes}

\newcommand\parens[1]{\left(#1\right)}
\newcommand\empt[2]{\(#1_{#2}\)}

\begin{document}

\maketitle

\begin{abstract}
  We show that both an LSTM and a unitary-evolution recurrent neural
  network (URN) can achieve encouraging accuracy on two types of
  syntactic patterns: context-free  long distance agreement, and mildly context-sensitive 
  cross serial dependencies. This work extends recent experiments on deeply 
  nested context-free long distance dependencies, with similar results. 
  URNs differ from LSTMs  in that they avoid non-linear activation functions, and 
  they apply matrix multiplication to word embeddings encoded as unitary matrices. This
  permits them to retain all information in the processing of an input
  string over arbitrary distances. It also causes them to satisfy strict
  compositionality. URNs constitute a significant advance in the search 
  for explainable models in deep learning applied to NLP.
\end{abstract}

\section{Introduction}

\cite{coecke_mathematical_2010,grefenstette_concrete_2011} proposed
end-to-end-trainable vector space semantic models based on the syntactic 
types of a pregroup grammar \cite{lambek_pregroup_2008}.
More recently, \cite{mcpheat_categorical_2021} construct a vector space semantics
with a modality, within a Lambek calculus-based type logical grammar.\footnote{\cite{wijnholds_representation_2020} 
  use the syntactic type representations of a Combinatory Categorial Grammar 
  \cite{Steedman2000} to train a deep neural network to learn word representations.}

A significant inspiration for much of this work is the category theory
within which  quantum mechanics is formulated. The types of the 
Lambek calculus grammar generate a syntactic structure that is
interpreted through the category of matrices. The set of such
matrices constitutes a semantic representation for lexical
items. They are anchored in distributional word vectors that 
correspond to the word embeddings of contemporary deep neural networks.
The values of these matrices can be set to optimise a variety
of NLP objectives, such as the evaluation of semantic distance and relatedness
among sentences. Following this template, one obtains what the authors
call a \emph{compositional vector space semantics}. 
The compositionality of these semantic representations comes 
from the type system of the grammar, which is assumed as 
given.

In this paper, we consider a model of compositional vector
\emph{syntax}, based on premises related to those assumed by the work just
described. The main similarity with this work is that the word representations 
are unitary matrices, which can be trained end-to-end. The main difference
is that our model does not rely on the types of an existing syntactic 
representation. Instead it learns syntactic structure directly from the data. 

The only algebraic structure that our model invokes is that of
sequences--- mathematically, a free monoid over input symbols. The
relevant compositional principle specifies that the representation
\(⟦w⟧\) of an input sequence \(w\) is a monoid homomorphism, defined using an
associative combination operator \((·)\). This principle requires that
the following equation holds for any two sub-sequences \(w_1\) and \(w_2\):
\begin{equation}
⟦w_1 w_2⟧ = ⟦w_1⟧ · ⟦w_2⟧ \label{eq:monoid-homo}
\end{equation}
Hence the representation of a concatenation is the
combination of representations with the operator
\((·)\). Additionally, the representation of the empty string must be
the unit of \((·)\).

In this paper, we consider the unitary-evolution recurrent neural
network (URN), first suggested by \cite{arjovsky_unitary_2016}.  This
is a kind of RNN where the function applied to the hidden state vector
is a unitary transformation. We analyse the capability of the URN to
model two syntactic patterns: a context free structure and a mildly context
sensitive one. We do this by looking at synthetic languages, which
allows us to abstract from the noise that pervades natural language
corpus data. We compare the syntactic capabilities of the URN
with the LSTM, the dominant architecture for current RNNs.

We recap the definitions of the models and study the theoretical
properties of the URN in \cref{sec:models}. Our experiments are
described in \cref{sec:cs-deps-ex,sec:dyck-ex}. We discuss the results
in \cref{sec:discussion}, and related work in \cref{sec:related-work},
before summarising our conclusions in \cref{sec:conclusion}.

\section{Models}
\label{sec:models}

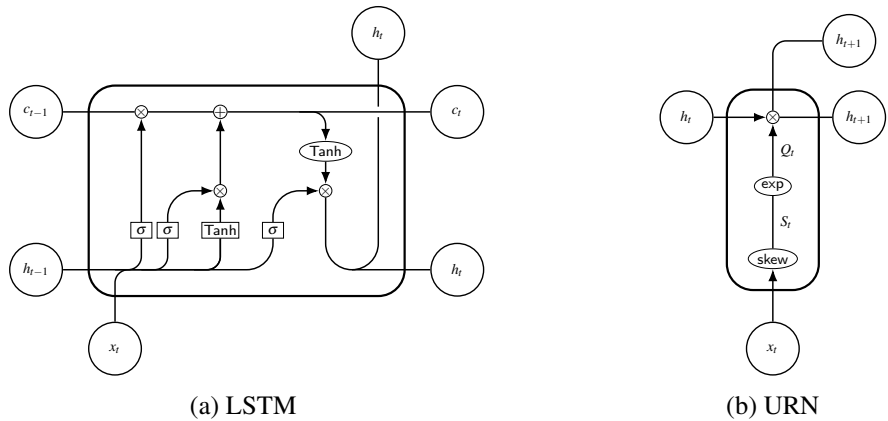
\begin{figure}
  \centering

\subcaptionbox{LSTM}{
  \scalebox{0.7}{
\begin{tikzpicture}[
    font=\sf \scriptsize,
    >=LaTeX,
    cell/.style={
        rectangle, 
        rounded corners=5mm, 
        draw,
        very thick,
        },
    operator/.style={
        circle,
        draw,
        inner sep=-0.5pt,
        minimum height =.2cm,
        },
    function/.style={
        ellipse,
        draw,
        inner sep=1pt
        },
    ct/.style={
        circle,
        draw,
        line width = .75pt,
        minimum width=1cm,
        inner sep=1pt,
        },
    gt/.style={
        rectangle,
        draw,
        minimum width=4mm,
        minimum height=3mm,
        inner sep=1pt
        },
    mylabel/.style={
        font=\scriptsize\sffamily
        },
    ArrowC1/.style={
        rounded corners=.25cm,
        thick,
        },
    ArrowC2/.style={
        rounded corners=.5cm,
        thick,
        }
    ]

    \node [cell, minimum height =4cm, minimum width=6cm] at (0,0){} ;

    \node [gt] (ibox1) at (-2,-0.75) {$\sigma$};
    \node [gt] (ibox2) at (-1.5,-0.75) {$\sigma$};
    \node [gt] (ibox3) at (-0.5,-0.75) {Tanh};
    \node [gt] (ibox4) at (0.5,-0.75) {$\sigma$};

    \node [operator] (mux1) at (-2,1.5) {$\times$};
    \node [operator] (add1) at (-0.5,1.5) {+};
    \node [operator] (mux2) at (-0.5,0) {$\times$};
    \node [operator] (mux3) at (1.5,0) {$\times$};
    \node [function] (func1) at (1.5,0.75) {Tanh};

    \node[ct] (c) at (-4,1.5) {\empt{c}{t-1}};
    \node[ct] (h) at (-4,-1.5) {\empt{h}{t-1}};
    \node[ct] (x) at (-2.5,-3) {\empt{x}{t}};

    \node[ct] (c2) at (4,1.5) {\empt{c}{t}};
    \node[ct] (h2) at (4,-1.5) {\empt{h}{t}};
    \node[ct] (x2) at (2.5,3) {\empt{h}{t}};

    \draw [ArrowC1] (c) -- (mux1) -- (add1) -- (c2);

    \draw [ArrowC2] (h) -| (ibox4);
    \draw [ArrowC1] (h -| ibox1)++(-0.5,0) -| (ibox1); 
    \draw [ArrowC1] (h -| ibox2)++(-0.5,0) -| (ibox2);
    \draw [ArrowC1] (h -| ibox3)++(-0.5,0) -| (ibox3);
    \draw [ArrowC1] (x) -- (x |- h)-| (ibox3);

    \draw [->, ArrowC2] (ibox1) -- (mux1);
    \draw [->, ArrowC2] (ibox2) |- (mux2);
    \draw [->, ArrowC2] (ibox3) -- (mux2);
    \draw [->, ArrowC2] (ibox4) |- (mux3);
    \draw [->, ArrowC2] (mux2) -- (add1);
    \draw [->, ArrowC1] (add1 -| func1)++(-0.5,0) -| (func1);
    \draw [->, ArrowC2] (func1) -- (mux3);

    \draw [-, ArrowC2] (mux3) |- (h2);
    \draw (c2 -| x2) ++(0,-0.1) coordinate (i1);
    \draw [-, ArrowC2] (h2 -| x2)++(-0.5,0) -| (i1);
    \draw [-, ArrowC2] (i1)++(0,0.2) -- (x2);

\end{tikzpicture}
}}
\hspace{2cm}
  \subcaptionbox{URN}{\scalebox{0.7}{
    \begin{tikzpicture}[
      font=\sf \scriptsize,
      >=LaTeX,
      cell/.style={
        rectangle, 
        rounded corners=5mm, 
        draw,
        very thick,
      },
      operator/.style={
        circle,
        draw,
        inner sep=-0.5pt,
        minimum height =.2cm,
      },
      function/.style={
        ellipse,
        draw,
        inner sep=1pt
      },
      ct/.style={
        circle,
        draw,
        line width = .75pt,
        minimum width=1cm,
        inner sep=1pt,
      },
      gt/.style={
        rectangle,
        draw,
        minimum width=4mm,
        minimum height=3mm,
        inner sep=1pt
      },
      mylabel/.style={
        font=\scriptsize\sffamily
      },
      ArrowC1/.style={
        rounded corners=.25cm,
        thick,
      }
      ]


      \node [operator] (mux) {$\times$};
      \node [function,below=of mux] (exp) {exp};
      \node [function,below=of exp] (antisym) {skew};

      \node [cell,fit=(mux) (antisym),inner sep=0.4cm]{} ;

      \node[ct,left=of mux]    (h) {\empt{h}{t}};
      \node[ct,below=of antisym] (x) {\empt{x}{t}};

      \node[ct,above right=of mux] (h2) {\empt{h}{t+1}};
      \node[ct,right=of mux] (x2) {\empt{h}{t+1}};

      \draw [->,ArrowC1] (x) -- (antisym);
      \path [ArrowC1] (antisym) edge node[right]{\(S_t\)} (exp);
      \path [->,ArrowC1] (exp) edge node[right]{\(Q_t\)} (mux);
      \draw [->,ArrowC1] (h) -- (mux);
      \draw [ArrowC1] (mux) -- (x2);
      \draw [ArrowC1] (mux) |- (h2);
    \end{tikzpicture}
  }}
  \caption{Schematic representation of models}
  \label{fig:models-schematic}
  
\end{figure}

We consider two generative models: an LSTM
\cite{hochreiter_long_1997} and an URN
\cite{arjovsky_unitary_2016}. We portray them schematically in
  \cref{fig:models-schematic}. All models are trained end-to-end as
  generative language models: we use cross-entropy loss for each
  symbol in the training set, which we sum for each position, up to
  the stop symbol. There is no task-specific supervised training. The
  training method is by stochastic gradient descent. More precisely,
  we use an Adam optimiser with a learning rate of \(0.001\), and a
  batch size of 512. The number of parameters for each model is listed
  in \cref{tab:num-par,tab:num-par-dyck}.

\subsection{LSTM}
We provide the definition of an LSTM here to specify which
version we are using, and to highlight the contrasts with URNs.
\begin{align*}
v_t        & = h_{t-1} ◇ x_t \\
f_t        & = σ(W_f v_t + b_f) \\
i_t        & = σ(W_i v_t + b_i) \\
o_t        & = σ(W_o v_t + b_i) \\
\tilde c_t & = σ(W_c v_t + b_c) \\
c_t        & = (f_t ⊙ c_{t-1}) + (i_t ⊙ \tilde c_t) \\
h_t        & = (o_t ⊙ \mathsf{tanh}(c_t))
\end{align*}
Here $σ$ refers to the sigmoid function and $(◇)$ is vector concatenation.
We apply dropout to the vectors $h_t$ and $ x_t$ for every timestep $t$. Predictions
are obtained by applying a projection layer to $h_t$, with softmax
activation. The input $x_t$ is obtained by an embedding layer.

\subsection{URN}
The URN, in the variant that we employ, has a much simpler definition.
\begin{align*}
  h_t        & = Q_t h_{t-1} \\
  S_t & = skew(x_t) \\
  Q_t & = e^{S_t}
\end{align*}
\(skew(x)\) is a function that takes a vector and produces a
skew-symmetric matrix by arranging the elements of \(x\) in a
triangular pattern. For example, with an input vector of size 3, we
have:
\[skew(x) = \parens{
    \begin{smallmatrix}
      0 & x_{0} & x_{1}  \\
      -x_{0} & 0 & x_{2} \\
      -x_{1} & -x_{2} & 0
    \end{smallmatrix}}\]
The upper triangle of $S_t$ is provided by the previous
layer (typically the word embedding layer), and its lower triangle is
its negated symmetric.  This setup ensures that $S_t$ is
anti-symmetric. This feature, together with the properties of matrix
exponential, insures that $Q_t$ is unitary. We call \(Q_t\) the
\emph{unitary embedding} of the input symbol at position \(t\).

\paragraph{Compositional property}

We can now show that the URN exhibits compositionality, in the sense
of \cref{eq:monoid-homo}.
In general, if we let \(⟦w[t]⟧ = Q_t\), then the output of the URN
model for any input sequence is:
\begin{align*}
  h_{n+1} & = Q_n × (Q_{n-1} × (⋯   (Q_0 × h_0))) \\
         &  = ⟦w[n]⟧ × (⟦w[{n-1}]⟧ × (⋯ × (⟦w[0]⟧ × h_0))) \\
         &  = (⟦w[n]⟧ × ⟦w[{n-1}]⟧ × ⋯ × ⟦w[0]⟧) × h_0 & \text{by associativity of matrix-vector product}\\
         &  = ⟦w[0] … w[n]⟧ × h_0
\end{align*}
With:
\begin{align*}
  ⟦w[t]⟧ = e^{skew(x_t)} \tag{atomic symbol} \\
  ⟦w_1 w_2⟧ = ⟦w_2⟧ × ⟦w_1⟧ \tag{concatenation}
\end{align*}
The above definition of \(⟦\_⟧\) satisfies
\cref{eq:monoid-homo}, with \(P · Q = Q × P\).\footnote{We leave it to the reader
  to check that this definition of the \((·)\) operator is associative.}
Because unitary matrices form a monoid under product, \(⟦w⟧\) can be
represented by a unitary matrix for any input string \(w\).

\paragraph{Long-distance properties of URN models}

There are two key properties of unitary interpretations that motivate
their use.  First, the absence of non-linear activation functions on
the recurrent path entails that they are not subject to exploding or
vanishing gradients, as \cite{arjovsky_unitary_2016} observe.  Second,
because the transformations involved in their processing of input are
always unitary, any difference in the start state is preserved in the
output state. For every unitary matrix \(Q\), we have
\begin{equation*}
  ⟨Qh,Qs⟩ = ⟨h,s⟩
\end{equation*}
Another manifestation of this property is that unitary matrices can
always be inverted, by transposition:\footnote{Additionally, one
  should take the complex conjugate when dealing with complex-valued
  matrices.}
\begin{equation*}
  Q^*Q = I
\end{equation*}
As a consequence, no information is lost through time steps. This
formal analysis suggests that the URN will be good at tasks which
require long-term memory.  We devote the remainder of this paper to
confirming experimentally that this property is observed for
context-free and mildly context-sensitive inputs, when training the
word embeddings by stochastic gradient descent (using the Adam 
optimiser).

\paragraph{Training regime}
We apply dropout to the matrix $S_t$ for every timestep $t$. The input
matrix $S_t$ has dimensions $n×n$, with $n$ being referred as the
number of units below. Predictions are obtained in the same way as for
the LSTM.

\section{Cross-Serial dependencies}
\label{sec:cs-deps-ex}

\cite{shieber_evidence_1985} demonstrated the non-context-free nature of
interleaved verb-object relations in Dutch and Swiss German. One of
Shieber's examples of an embedded verb-object crossing dependency in
Swiss German is given in example \ref{cross-serial} below.
\begin{lingex}
\item Jan sät das mer {\bf{d’chind}} em {\bf{\emph{Hans}}} es huus {\bf{lönd}} {\bf{\emph{hälfe}}} aastriiche\\
  \label{cross-serial}
  Jan said that we the children-ACC Hans-DAT the house-ACC let help paint\\ 
  {\emph{Jan said that we let the children help Hans paint the house.}}
\end{lingex}
Similar patterns have been observed in other languages. They can be expressed
by indexed grammars \cite{aho_indexed_1968,pulman_indexed_1985}, as well as a
variety of other Mildly Context-Sensitive grammar formalisms
\cite{Joshi&etal1990,stabler_varieties_2004}.

\cite{shieber_evidence_1985} observed that cross-serial dependency patterns
of case marked nouns and their corresponding verbs can be iterated in
this construction. The above pattern can be abstracted as a
set of $a^mb^nc^md^n$ structures, which together form a Mildly Context-Sensitive
language.


Formally, we consider the family of languages
$\Lang_k = \braces {a^mb^nc^md^n \mid m+n < k}$. Note that if
\(k < l\), then \(\Lang_k \subset \Lang_l\). The training set
consists of 51,200 strings picked uniformly from \({\Lang_{8}}\). The
test set contains 5,120 strings picked uniformly from \({\Lang_{10}}\).

We recall that the RNNs are trained as generative language
models. That is, assuming a sample string $w ∈ \Lang_{10}$, RNNs are
trained to predict the symbol $w_{i+1}$ given $w_0$ to $w_i$. Special
start and a stop symbols are added to the input strings, as is
standard.

This is to be contrasted with the testing procedure.
At test time, given the prefix $w_0…w_i$, a prediction of symbol
$w_{i+1}$ is deemed correct if it is a possible continuation for
$w_0…w_i$; that is, if $w_0…w_{i+1}$ is a prefix of some string in
\({\Lang_{10}}\). A set of predictions for a full string $x_0…x_k$ is classified
as correct, if all predictions are correct up to and including the
stop symbol. We report error rates for full strings only. This is
because when models make a mistake, it is typically for a single
symbol near the end of a string.

\begin{table}
    \centering
    \begin{tabular}[h]{rrr}
      Number of units & LSTM & URN \\ \hline
      32 & 7314& 5290 \\
      16 & 2738 & 1370 \\
      8  & 1218 & 370 \\
    \end{tabular}
    \caption{Number of parameters for cross-dependency pattern models}
    \label{tab:num-par}
  \end{table}

\subsection{Results}

Both RNNs struggle to generalise these patterns. They can model the
training data well, but produce incorrect patterns in some cases on
strings of any greater length than the samples in the training corpus.

\begin{figure}
  \centering
  \begin{subfigure}{7.5cm}
    \centering
    \begin{tikzpicture}[scale=0.9] 
      \begin{axis} 
        \addplot plot table[x=epoch,y=testloss]{data/lstm32.txt};  
        \addplot plot table[x=epoch,y=testloss]{data/lstm16.txt};  
        \addplot plot table[x=epoch,y=testloss]{data/lstm8.txt};   
        \addplot plot table[x=epoch,y=testloss]{data/urn32.txt};   
        \addplot plot table[x=epoch,y=testloss]{data/urn16.txt};   
        \addplot plot table[x=epoch,y=testloss]{data/urn8.txt};    
      \end{axis} 
    \end{tikzpicture}
    \caption{Test loss across epochs}
    \label{fig:test-loss}
  \end{subfigure}
~
  \begin{subfigure}{7.5cm}
    \centering
    \begin{tikzpicture}[scale=0.9] 
      \begin{axis}
        \addplot plot table[x=epoch,y expr=1-\thisrow{accuracy}]{data/lstm32.txt};    
        \addplot plot table[x=epoch,y expr=1-\thisrow{accuracy}]{data/lstm16.txt};    
        \addplot plot table[x=epoch,y expr=1-\thisrow{accuracy}]{data/lstm8.txt};     
        \addplot plot table[x=epoch,y expr=1-\thisrow{accuracy}]{data/urn32.txt};     
        \addplot plot table[x=epoch,y expr=1-\thisrow{accuracy}]{data/urn16.txt};     
        \addplot plot table[x=epoch,y expr=1-\thisrow{accuracy}]{data/urn8.txt};      
      \end{axis} 
    \end{tikzpicture}
    \caption{Error rate across epochs}
    \label{fig:err-rate}
  \end{subfigure}

    \quad\vspace{1cm}

  \begin{subfigure}{7.5cm}
  \centering
  \begin{tikzpicture}[scale=0.9] 
    \begin{axis}[x dir=reverse,legend style={anchor=south west,at={(0.05,0.05)}}]
      \addplot plot table[x=trainloss,y expr=1-\thisrow{accuracy}]{data/lstm32.txt};  \addlegendentry{32-unit LSTM}
      \addplot plot table[x=trainloss,y expr=1-\thisrow{accuracy}]{data/lstm16.txt};  \addlegendentry{16-unit LSTM}
      \addplot plot table[x=trainloss,y expr=1-\thisrow{accuracy}]{data/lstm8.txt};   \addlegendentry{8-unit LSTM}
      \addplot plot table[x=trainloss,y expr=1-\thisrow{accuracy}]{data/urn32.txt};   \addlegendentry{32-unit URN}
      \addplot plot table[x=trainloss,y expr=1-\thisrow{accuracy}]{data/urn16.txt};   \addlegendentry{16-unit URN}
      \addplot plot table[x=trainloss,y expr=1-\thisrow{accuracy}]{data/urn8.txt};    \addlegendentry{8-unit URN}
    \end{axis} 
  \end{tikzpicture}
  \caption{Error rate against training loss.}
  \label{fig:training-loss-error}
\end{subfigure}
  ~
  \begin{subfigure}{7.5cm}
    \centering
    \begin{tikzpicture}[scale=0.9] 
      \begin{axis}[xmax=10]
        \addplot plot table[x=p,y expr=1-\thisrow{acc}]{npm/lstm32.txt};   
        \addplot plot table[x=p,y expr=1-\thisrow{acc}]{npm/lstm16.txt};   
        \addplot plot table[x=p,y expr=1-\thisrow{acc}]{npm/lstm8.txt};    
        \addplot plot table[x=p,y expr=1-\thisrow{acc}]{npm/urn32.txt};    
        \addplot plot table[x=p,y expr=1-\thisrow{acc}]{npm/urn16.txt};    
        \addplot plot table[x=p,y expr=1-\thisrow{acc}]{npm/urn8.txt};     
      \end{axis} 
    \end{tikzpicture}
    \caption{Error rate against \(n+m\), at epoch 100}
    \label{fig:dyck-err-rate-by-attractors}
  \end{subfigure}
  
  \caption{Cross-Serial dependencies results for various models}
  \label{fig:cross-serial-results}
\end{figure}
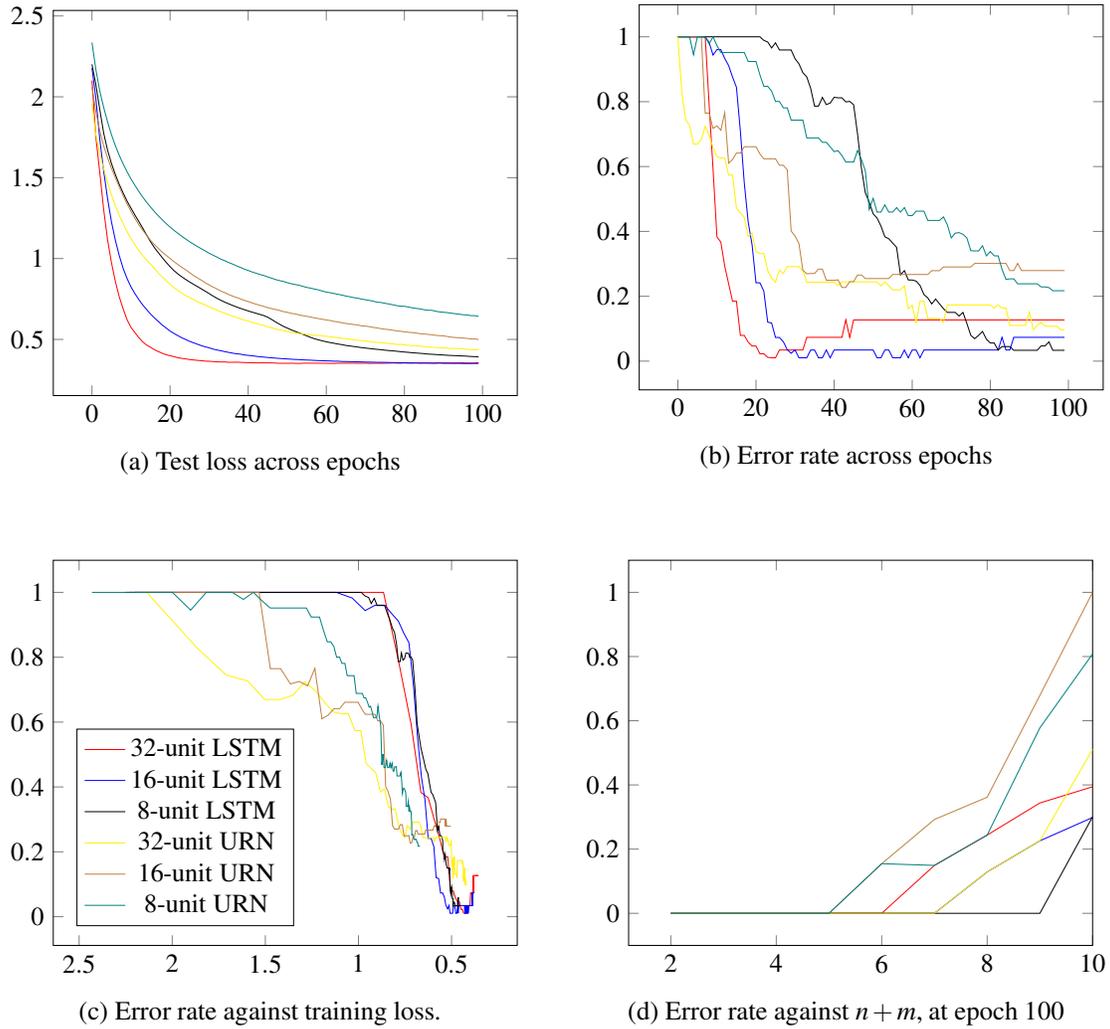

We report four sets of results.
The first set (\cref{fig:test-loss}) is the cross-entropy loss for the
\emph{test set} obtained by each model across training epochs. Low
losses indicate that, on a per-character basis, the models reproduce
the \emph{exact} strings in the test set, rather than making correct
predictions, as defined above. We see that LSTM models generally make
better guesses than URNs, across the board.

The second set (\cref{fig:err-rate}) is the error rate over number of
epochs, for each tested model. The best models can achieve less then
ten percent error rate on average on the test set. However, the LSTM
models with a larger number of units exhibit overfitting.

In the third set we show the error rate for a given training loss in
\cref{fig:training-loss-error}. The corresponding ratio is a measure
of a model's capacity for correct predictions of a given quality of
approximation of the training set. It can be taken as an indication of
the model's bias for this task, relative to generative language
modelling. The URN models tend to achieve lower error
rates for this task, even though they do less well from a generative
language modelling perspective. For instance, the 32-unit URN is able
to obtain an error rate below 0.4 with a training loss as high as
1. In general, the URNs provide a smoother decrease in error
rate as they learn the language.  In contrast, the LSTM models exhibit
a sharp drop in error rate around 0.7 training loss.

Finally, we show the error rate broken down by length of pattern
(reported as \(n+m\)). We see here that the LSTMs tend to do better
overall than the URNs for lengths unseen in the training
data. However, the LSTMs do worse when the number of units increases,
while the URNs do better as that number increases, thanks to a lack of
overfitting.

\section{Generalised Dyck Languages}
\label{sec:dyck-ex}

In the next experiment, we evaluate the long-distance modelling
capabilities of an RNN for a context-free language. As before, we do
it in a way that abstracts away from the noise of natural language, by
constructing synthetic data. Following \cite{bernardy_can_2018}, we
use a (generalised) Dyck language. This language is composed solely
of matching parenthesis pairs. So the strings ``\verb:{([])}<>:'' and
``\verb:{()[<>]}:'' are part of the language, while ``\verb!<)!'' is
not.

Formally, we use the language \(\mathcal D\) defined as the set of strings generated
by the following context-free rules: $E ::= ε; E ::= EE; E ::= oEc$,
where $(o,c)$ stands for a pair of matching parenthesis pairs.  In all
of our tests, we use $5$ types of pairs (corresponding, for
example, to the pairs \verb!()! , \verb![]!, \verb!{}!, \verb!<>!  and
\texttt{`'}.)

The aim of the task is to predict the correct type of \emph{closing}
parenthesis at every point in a string. It should be noted that this
experiment is an idealised version of the agreement task proposed by
\cite{linzen_assessing_2016}. The opening parenthesis plays
the role of a word (say a noun) which governs a feature of a
subsequent word (say the number of a verb), represented by the closing
parenthesis. Matching of parentheses corresponds to agreement.
\cite{linzen_assessing_2016} point out that sustaining accuracy over
long distances requires that the model have knowledge of hierarchical
syntactic structure. If an RNN captures the long-distance
dependencies involved in agreement relations, it cannot rely solely
on the nearby governing symbols. In particular, the accuracy must be
sustained as the number \emph{attractors} increases. For our
experiment, an attractor is defined as an opening parenthesis
occurring within a matching pair, but of the wrong kind. For instance,
in ``\verb:{()}:'', the parenthesis ``\verb:(:'' is an attractor.

To generate a string with $N$ matching pairs, we perform a random walk
between opposite corners of a square grid of width and height $N$,
such that one is not allowed to cross the diagonal. When not
restricted by the boundary, a step can be taken either along the $x$
or $y$ axis with equal probability.  A step along the $x$ axis
corresponds to opening a parenthesis, and one along the $y$ axis
involves closing one. The type of parenthesis pair is chosen
randomly and uniformly.

In this task, we use strings with a length of exactly 20
characters. We train on 102400 and test on 5120 random strings. In the
previous experiment we varied the string length between training and
testing.  In this experiment, we vary the \emph{nesting depth}, from 3
to 9.  For this purpose, we define the depth of the string is the
maximum nesting level reached within it. For instance ``\verb:[{}]:''
has depth 2, while ``\verb:{([()]<>)}:'' has depth 4.

As in the first experiment, for the training phase the RNNs are treated
as generative language models, applying a cross-entropy loss function
at each position in the string.  At test time, we evaluate the model's
ability to predict the right kind of closing parenthesis at each
point. We ignore predictions regarding opening parentheses, because
they are always acceptable for the language.

Training is performed with a learning rate of 0.01, and a dropout
rate of \(ρ=0.05\), for 100 epochs.

\subsection{Results}

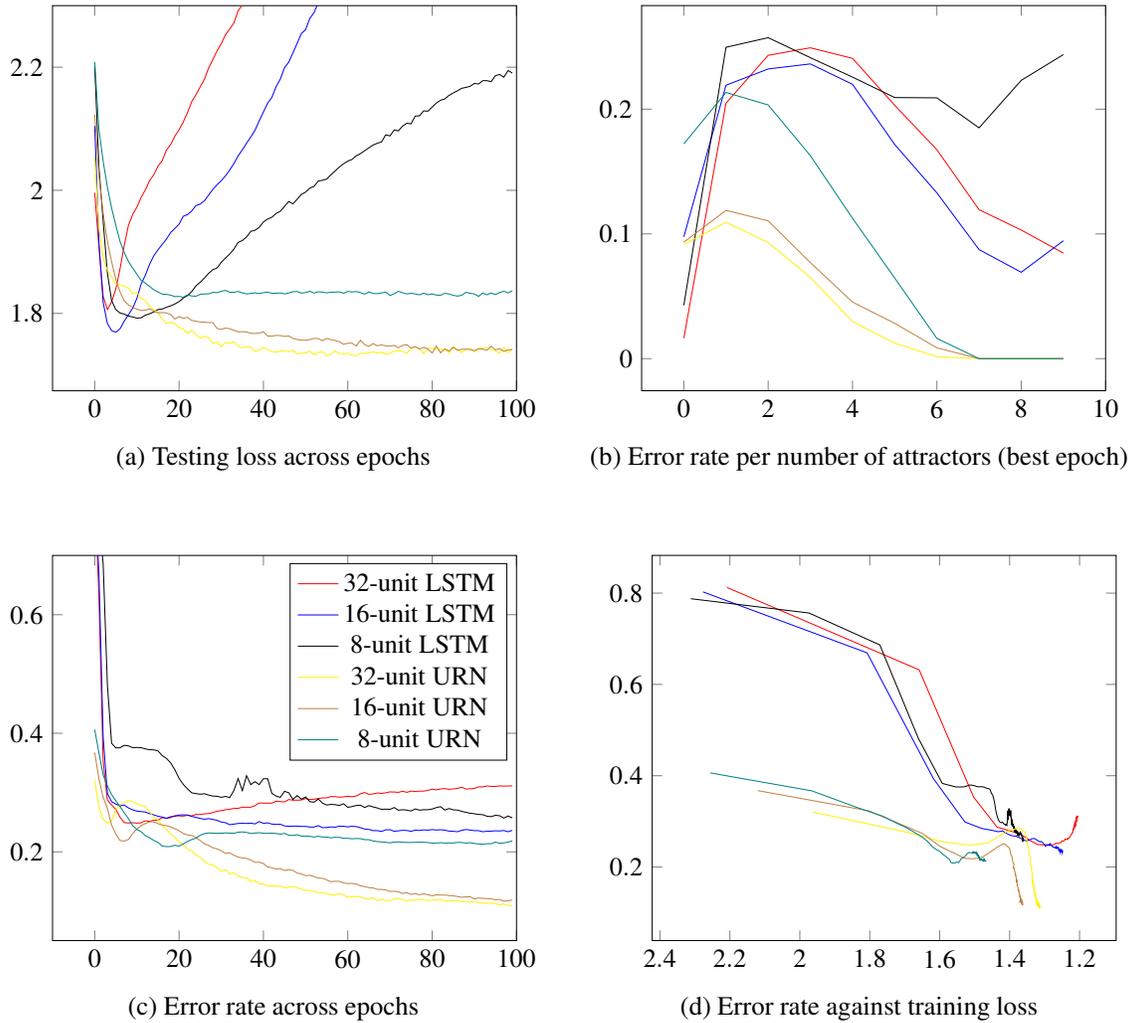
\begin{figure}\centering
  \begin{subfigure}{7.5cm}
    \centering
    \begin{tikzpicture}[scale=0.9] 
      \begin{axis}[xmax=100,ymax=2.3]
        \addplot plot table[x=epoch,y=testloss]{dyck/lstm32.txt};    
        \addplot plot table[x=epoch,y=testloss]{dyck/lstm16.txt};    
        \addplot plot table[x=epoch,y=testloss]{dyck/lstm8.txt};     
        \addplot plot table[x=epoch,y=testloss]{dyck/urn32.txt};     
        \addplot plot table[x=epoch,y=testloss]{dyck/urn16.txt};     
        \addplot plot table[x=epoch,y=testloss]{dyck/urn8.txt};      
      \end{axis} 
    \end{tikzpicture}
    \caption{Testing loss across epochs}
    \label{fig:dyck-test-loss}
  \end{subfigure}
  ~
  \begin{subfigure}{7.5cm}
    \centering
    \begin{tikzpicture}[scale=0.9] 
      \begin{axis}[xmax=10]
        \addplot plot table[x=metric,y expr=1-\thisrow{accuracy}]{dyck-err/lstm32.txt};   
        \addplot plot table[x=metric,y expr=1-\thisrow{accuracy}]{dyck-err/lstm16.txt};   
        \addplot plot table[x=metric,y expr=1-\thisrow{accuracy}]{dyck-err/lstm8.txt};    
        \addplot plot table[x=metric,y expr=1-\thisrow{accuracy}]{dyck-err/urn32.txt};    
        \addplot plot table[x=metric,y expr=1-\thisrow{accuracy}]{dyck-err/urn16.txt};    
        \addplot plot table[x=metric,y expr=1-\thisrow{accuracy}]{dyck-err/urn8.txt};     
      \end{axis} 
    \end{tikzpicture}
    \caption{Error rate per number of attractors (best epoch)}
    \label{fig:dyck-err-rate-by-attractors}
  \end{subfigure}

  \quad\vspace{1cm}

  \begin{subfigure}{7.5cm}
    \centering
    \begin{tikzpicture}[scale=0.9] 
      \begin{axis}[xmax=100,ymax=0.7]
        \addplot plot table[x=epoch,y=maxErrRate]{dyck/lstm32.txt};     \addlegendentry{32-unit LSTM}
        \addplot plot table[x=epoch,y=maxErrRate]{dyck/lstm16.txt};     \addlegendentry{16-unit LSTM}
        \addplot plot table[x=epoch,y=maxErrRate]{dyck/lstm8.txt};      \addlegendentry{8-unit LSTM}
        \addplot plot table[x=epoch,y=maxErrRate]{dyck/urn32.txt};      \addlegendentry{32-unit URN}
        \addplot plot table[x=epoch,y=maxErrRate]{dyck/urn16.txt};      \addlegendentry{16-unit URN}
        \addplot plot table[x=epoch,y=maxErrRate]{dyck/urn8.txt};       \addlegendentry{8-unit URN}
      \end{axis} 
    \end{tikzpicture}
    \caption{Error rate across epochs}
    \label{fig:dyck-err-rate-by-epoch}
  \end{subfigure}
  ~
  \begin{subfigure}{7.5cm}
    \centering
    \begin{tikzpicture}[scale=0.9] 
      \begin{axis}[x dir=reverse]
        \addplot plot table[x=trainloss,y=maxErrRate]{dyck/lstm32.txt};    
        \addplot plot table[x=trainloss,y=maxErrRate]{dyck/lstm16.txt};    
        \addplot plot table[x=trainloss,y=maxErrRate]{dyck/lstm8.txt};     
        \addplot plot table[x=trainloss,y=maxErrRate]{dyck/urn32.txt};     
        \addplot plot table[x=trainloss,y=maxErrRate]{dyck/urn16.txt};     
        \addplot plot table[x=trainloss,y=maxErrRate]{dyck/urn8.txt};      
      \end{axis} 
    \end{tikzpicture}
    \caption{Error rate against training loss}
    \label{fig:dyck-err-rate-by-train-loss}
  \end{subfigure}
  \caption{Dyck language experiment results. When reporting error
    rates as training progresses, we use the maximum error rate across
    number of attractors.}
\end{figure}

We report four sets of results.
The first set (\cref{fig:dyck-test-loss}) is the cross-entropy loss
for the \emph{test set} achieved by each model across training
epochs. Low losses indicate that, on a per-character basis, the models
reproduce the \emph{exact} strings in the test set. These losses
cannot drop to zero because it is always valid to predict an opening
parenthesis.  As in the cross-serial task, we observe that LSTM
models make better guesses than URN, at least for a similar number of
units. However, we see that the training of the URN is uniformly monotonous,
while the LSTM can sometimes become worse for a few epochs before
converging. In fact, for 8 units, the LSTM exhibits overfitting. The
test loss increases slowly after epoch 30.

To analyse the performance of each model on the task, we break down
the error rate by number of attractors
(\cref{fig:dyck-err-rate-by-attractors}). The URN models are weakest
for a low number of attractors, and they achieve near excellent accuracy
for a large number of attractors. According to
\cite{linzen_assessing_2016}, this suggests that the models are
highly successful in learning hierarchical structure. A high numbers
of attractors corresponds to outer pairs, while a low number of
attractors corresponds to inner pairs. In sum, in inner positions the
URN suffers from some confusion. This confusion decreases as the
number of units increases.

The LSTM is able to predict outer pairs rather well (but nowhere near
as successfully as the URN models). It reaches almost perfect
accuracy for adjacent pairs, with zero attractors, such as
\texttt{[]}.  It does worst on pairs with 3 to 4 attractors. The
LSTM is good at making a prediction which depends only on the previous
symbol. LSTM models with a larger number of units are also good at
making a prediction for a pair which encloses the whole
string. This indicates that it is fairly limited in its
ability to capture hierarchical structures over long distances, even
though it does much better than the majority class baseline, which
stands at an 80\% error rate.

In what follows, we will consider only the the \emph{maximum} error
rate for any given number of attractors, for varying epochs. That is,
we report the peak value from the previous graph as training
progresses. Using this metric, a URN performs consistently better
than an LSTM with the same number of units (and a similar number of
parameters). It does so consistently across the training period
(\cref{fig:dyck-err-rate-by-epoch}). However, we note that every model
is capable of generalising to deeper nesting levels to some extent,
with an accuracy well above a majority class baseline (with an
error rate of 80\%). The URN models beat the majority class baseline
within the first epoch, while the LSTM needs a couple of epochs to do
so.

Finally we report the error rate against training loss
(\cref{fig:dyck-err-rate-by-train-loss}), as we did for the
cross-serial dependency task. Again, we do not report the average
error rate, but rather the \emph{maximum} error rate for any given
number of attractors.  Here too, the relationship between error-rate
and training loss corresponds to the bias of the model for the task at
hand, compared to a generative language model task. We observe that
the URN models outperform the LSTM models across the board.

\begin{table}
    \centering
    \begin{tabular}[h]{rrr}
      Number of units & LSTM & URN \\ \hline
      32 & 6300 & 6348 \\
      16 & 2204 & 1644 \\
      8  & 924  & 444 \\
    \end{tabular}
    \caption{Number of parameters Dyck language models}
    \label{tab:num-par-dyck}
  \end{table}

\section{Discussion}
\label{sec:discussion}

In summary, both the URN and the LSTM perform reasonably well on our
experiments.  On the cross-serial dependency task, both architectures
can model the training data, but they do not generalise perfectly
to extended sequences. Still on both tasks, we observe that the URN
exhibits a bias for the task, rather than for pure generative
language model predictions. Additionally, the URN appears to be less
prone to overfitting, on both tasks.

These experiments are significant for at least two reasons. We believe
that the URN is the first RNN capable of recognising hierarchical
syntactic structures of the sort that characterise natural language
syntax.  While the LSTM can capture the patterns appearing in the
training set, the URN displays better generalisation capabilities
than the LSTM, in addition to being mathematically tractable. A
particularly attractive result, is that URN models achieve high
accuracy with less than a couple of thousand parameters.

Second, our experiments illustrate the effectiveness of URNs as
devices for tracking and predicting complex dependency relations, over
long strings, in a fully compositional way. Unlike LSTMs, they do not
suffer from opaqueness due to non-linear activation functions. They do
not make use of such functions, and so their behaviour is, in
principle, amenable to analysis using standard tools from linear
algebra.  They are not blackbox processing devices that require
indirect methods of analysis and assessment, as is the case with most
other deep neural networks.

\section{Related Work}
\label{sec:related-work}

\subsection{Long-distance agreement}

The capacity of Recurrent Neural Networks (RNNs), particularly LSTMs,
to identify context-free long distance dependencies has been widely
discussed in the NLP and cognitive science
literature~\cite{elman_finding_1990,linzen_assessing_2016,bernardy_using_2017,bernardy_can_2018,sennhauser_evaluating_2018,gulordava_colorless_2018,lappin_deep_2021}.
These discussions have considered dependency patterns in both
artificial systems, particularly Dyck languages, and in natural
languages, with subject-verb agreement providing a paradigm case.

\cite{elman_distributed_1991} already observed that it is useful to
experiment with artificial systems to filter out the noise of real
world natural language data.

\cite{bernardy_can_2018} tested the ability of the LSTM to predict
closing parenthesis types in a Dyck language. The results are
qualitatively similar. In both cases the LSTM makes the worse
predictions for a moderate number of attractors.  However he reports
worse results than us, despite using an LSTM with more units. We
attribute this difference to a better implementation of the LSTM. A
less perspicuous application of dropouts is a likely factor in the
poorer performance of \cite{bernardy_can_2018}'s LSTM
(\cite{bernardy_can_2018}). \cite{sennhauser_evaluating_2018}
performed a set of experiments with the same goal, and reported
results compatible with those of \cite{bernardy_can_2018}.

While LSTMs (and GRUs) exhibit a certain capacity to generalise to
deeper nesting, their accuracy declines in relation to nesting depth.
This is also the case with their handling of natural language
agreement. Other experimental work has illustrated this
effect~\cite{hewitt_rnns_2020,sennhauser_evaluating_2018}.  Similar
conclusions are observed for generative self-attention
architectures~\cite{yu_learning_2019}. Significantly, recent work has
indicated that non-generative self-attention architectures, in the
style of BERT, simply fail at this
task~\cite{bernardy_can_2021}. This suggests that sequential
processing is required to solve it.

By contrast URNs achieve excellent performance on this task, without
any decline in relation to either nesting depth, or number of
attractors. In recent work \cite{bernardy_neural_2022} provide an
explanation as to why this is the case, for a version of the URN
restricted to unitary matrices acting on 3 hyperplanes. They show
that the learned unitary embeddings for matching parentheses are
nearly the inverses of each other: \(⟦o\,c⟧ = ⟦c⟧×⟦o⟧ ≈ I\) for every pair of
matching parentheses \((o,c)\).  Such an analysis illustrates the 
role of compositionality in the performance of URNs on an NLP 
task.

\subsection{Cross-serial patterns}

\cite{kirov_processing_2012} study both nested and cross serial
dependencies with a Simple Recurrent Network (SRN).  As far as we are
aware, our experiment is the first application of both LSTMs and URNs
to cross serial dependency relations. While both achieve good
accuracy on the cross serial patterns, the URN offers significant
advantages in simplicity and transparency of architecture. It also
displays enhanced stability in learning, and power of structural
generalisation, relative to loss in training data.

\subsection{Unitary-Evolution Recurrent Networks}

\label{sec:unitary-evolution}

\cite{arjovsky_unitary_2016} propose Unitary-Evolution recurrent
networks to solve the problem of exploding and vanishing gradients,
caused by the presence of non-linear activation functions.  Despite
this, \cite{arjovsky_unitary_2016} suggest adding ReLU
activation between time-steps, unlike URNs. We are
primarily concerned with the structure of the underlying unitary
embeddings. The connection between the two lines of work is that if an
RNN suffers exploding/vanishing gradients, it cannot track long distance
dependencies.

Additionally, \cite{arjovsky_unitary_2016} use a complicated function to
transform word representations into unitary matrices. We use a simpler
method (exponential of anti-symmetric matrix), previously applied by
\cite{hyland_learning_2017} for deep learning models. This method has
performed well for the tasks discussed here. Because we use a fully
general matrix exponential implementation, our model is
computationally more expensive than others~\cite{arjovsky_unitary_2016,jing_tunable_2017}.  
When testing the unitary matrix encodings of
\cite{jing_tunable_2017} and \cite{arjovsky_unitary_2016}, we obtained
much worse results for our experiments. This may be because we do not
include ReLU activation, while they do. Another option for
enforcing the unitary character of matrices is to let
back-propagation update the unitary matrices arbitrarily
\(n \times n\), and project them onto the unitary space
periodically \cite{wisdom_full-capacity_2016,kiani_projunn_2022}.

\paragraph{Tensor Recurrent Neural Networks}

\cite{sutskever_generating_2011} describe what they call a “tensor
recurrent neural network” in which the transition matrix is determined
by each input symbol. This design appears to be similar to
URNs. However, unlike URNs, they use non-linear activation functions,
and so they inherit the complications that these functions produce.

\section{Conclusions}
\label{sec:conclusion}

The fact that URNs achieve good precision in predicting cross serial
dependencies and generalise appropriately for nested patterns suggest
that they are suitable for recognition of complex syntactic structures
of the sort that are challenging for other neural networks.

Our experiments show that URNs are biased towards predicting the
patterns found in context-free and mildly context-sensitive languages,
even when trained as generative language models. The fact that they
satisfy strict compositionality offers an important advance in the
search for explainable AI systems in deep learning models.

The move to powerful bidirectional transformers, like BERT, has
produced enhanced performance in a variety of NLP and other AI tasks.
This has been achieved at the expense of formal grounding and
computational transparency. It is even less obvious why such models
perform as well as they do on some tasks, and poorly on others, than
is the case for LSTMs. By contrast, URNs offer simple, light weight
deep neural networks whose operation is fully open to inspection and
understanding at each point in the processing regime. They can model
complex syntactic structures, in the sense that they can reproduce an
extensive training set of (artificial) data containing such
patterns. For cross-serial dependency patterns, they do not generalise
very well, but for hierarchical patterns they display impressive
generalisation capabilities.

URNs provide a principled solution to the problem of syntactic
compositionality.  They resolve the question of how to generate the
composite values of input arguments in a principled and
straightforward way. This is not the case for LSTMs, because the
combination of two cells cannot be expressed as a single cell. By
contrast, every URN cell applies matrix multiplication to its
constituents, and so the composition of the effects of two cells is
simply a matrix product.

URNs are worth exploring further as models of learning and
representation that bear some correspondence to human processing. 
In future work we will be studying the application of URNs to other
cognitively interesting NLP tasks. We are particularly interested in
examining possible parallels between the ways in which URNs handle
linguistic information, and human understanding of natural language
meaning and structure.

Finally, we observe that the unitary matrices through which URNs
compute output values from input arguments are identical to the gates
of quantum logic. This suggests the intriguing possibility of
implementing these models as quantum circuits. At some point in the
future, this may facilitate training these models on large amounts of
data, and efficiently generating results for tasks that are currently
beyond the resources of conventional computational systems.

In sum, we see the research that we report here as extending and
modifying some of the leading ideas in the foundational work of
\cite{coecke_mathematical_2010,grefenstette_concrete_2011}.  They
provided a system for handling computational semantics compositionally
with structures that map onto the matrices of quantum circuits.  We
offer a model for learning syntactic structure, and for processing in
general, that is strictly compositional. It uses some of the same core
methods as the earlier work.  In future research we will attempt to
apply our model to NLP tasks involving semantic interpretation.

\section*{Acknowledgements}

The research reported in this paper was supported by grant 2014-39
from the Swedish Research Council, which funds the Centre for
Linguistic Theory and Studies in Probability (CLASP) in the Department
of Philosophy, Linguistics, and Theory of Science at the University of
Gothenburg. We presented some of the main ideas of this paper to the
CLASP Seminar, in December 2021, and to the Cognitive Science Seminar of
the School of Electronic Engineering and Computer Science, Queen Mary
University of London, in February 2022.
We are grateful to the audiences of these two events for useful
discussion and feedback.

\bibliographystyle{eptcs}
\bibliography{jp,local}
\end{document}